\documentclass[conference]{IEEEtran}

\IEEEoverridecommandlockouts
\usepackage{cite}
\usepackage{amsmath,amssymb,amsfonts}
\usepackage{algorithmic}
\usepackage{graphicx}
\usepackage{textcomp}
\usepackage{xcolor}
\usepackage{subcaption}
\usepackage[draft]{hyperref} % Disable clickable links
\usepackage[letterpaper, top=0.75in, bottom=1.05in, left=0.625in, right=0.625in]{geometry}

\def\BibTeX{{\rm B\kern-.05em{\sc i\kern-.025em b}\kern-.08em
    T\kern-.1667em\lower.7ex\hbox{E}\kern-.125emX}}
\newcommand{\im}[1]{\ensuremath{#1}}
\newcommand{\kw}[1]{\im{\mathtt{#1}}}
\newcommand{\THESYSTEM}{\kw{PART}}    
\begin{document}

\title{ Multimodal  Injury Risk Prediction in Tennis  \\
 }

\author{\IEEEauthorblockN{Francisco Erramuspe Alvarez, Shobharani Polasa, Weihao Qu, Jay Wang and Ling Zheng }
\IEEEauthorblockA{\textit{Department of Computer Science and Software Engineering} \\
\textit{Monmouth University}\\
West Long Branch, USA \\
\{s1365567,s1365603, wqu, jwang, lzheng\}@monmouth.edu}
}
\newcommand{\wq}[1]{{#1}}

\maketitle
\begingroup
\renewcommand{\thefootnote}{}
\footnotetext{\textcopyright~2025 IEEE. Personal use of this material is permitted. Permission from IEEE must be obtained for all other uses, in any current or future media, including reprinting/republishing this material for advertising or promotional purposes, creating new collective works, for resale or redistribution to servers or lists, or reuse of any copyrighted component of this work in other works. Accepted author manuscript. Published in the 2025 IEEE 5th International Conference on Human-Machine Systems (ICHMS), pp. 28--34. DOI: 10.1109/ICHMS61085.2025.11154160.}
\addtocounter{footnote}{-1}
\endgroup

\begin{abstract}

Machine learning has had a significant positive impact on the prediction of athlete performance and injury risk. Most works in this field rely on subjective observations and expert assessments, which restrict their effectiveness. In sports like soccer, basketball, and wrestling, some studies attempt to address this challenge by integrating data from alternative sources, such as readings from wearable devices, alongside traditional subjective observations and expert assessments to enhance accuracy. However, similar research in tennis remains largely unexplored.

In this paper, we propose a multimodal Predictive Athlete Readiness framework for Tennis ({\THESYSTEM}) to assess both performance and injury risk in tennis players. By leveraging machine learning and deep learning techniques, {\THESYSTEM} processes multiple sources of data collected from nine collegiate tennis players, including physiological metrics, training and match data, sleep data from wearable devices, self-reported information via daily questionnaires, jump assessments, and motion analysis from match play videos. {\THESYSTEM} captures four characteristics of tennis players: the overall wellness, injury risk, physical capability, and playing style. By integrating these four characteristics by supervised learning, it is capable of providing a holistic assessment of the tennis athlete's condition, along with advanced forecasts of specific body areas at risk such as the upper body (e.g., elbows) or lower body (e.g., knees).
Our evaluation, conducted with data from nine collegiate tennis players, shows that PART achieves strong performance in predicting both overall wellness and injury risk. Additionally, our framework also shows promise for recreational tennis players, who often suffer from injuries due to incorrect playing techniques.

\end{abstract}

\section{Introduction}
In professional sports, optimizing athletic performance while minimizing the risk of injury requires a comprehensive understanding of the many factors that impact an athlete's well-being, including physiological, psychological, training, and lifestyle factors.
A popular approach to studying how these factors affect athletes’ physical health and injury risk involves applying machine learning to predict performance outcomes and potential injuries. Sports performance and injury prediction has been actively researched across sports like basketball, soccer, wrestling, and tennis. 
Recent studies in this area have highlighted key factors, such as workout schedules, player characteristics, injury history, and workload, as critical influences on injury risk~\cite{myers2020acute,moreno2021association,liu2023sports}. 

 Among studies on performance and injury prediction in various sports, research focused on tennis holds particular significance:
 (1) Tennis enjoys a unique and prestigious place in global sports, recognized as one of the most popular individual sports worldwide and attracting a dedicated fan base across all age groups.
(2) Tennis is complex, requiring a blend of athleticism, strategic thinking, and mental resilience, which makes performance and injury prediction especially challenging.
(3) Tennis-related injuries are common, often arising from overuse and repetitive motions~\cite{tennisinjury}. 
(4) A significant number of casual tennis players suffer from injuries caused by incorrect playing techniques, with well-known examples including tennis elbow and tennis knee~\cite{recreationaltennis}. This highlights the importance of injury prediction, especially for recreational players looking to prevent such issues.

However, current machine-learning-based performance and injury prediction in tennis still relies on official statistical game data, subjective observations, and expert assessments. This leads to a common challenge in existing research in sports performance and injury prediction: many models rely heavily on subjective observations and expert assessments~\cite{huang2022novel}, limiting their ability to capture the complex interplay of various factors. To this end, researchers are increasingly turning to diverse data sources that are better suited to machine learning models, aiming to improve the accuracy of performance and injury risk predictions.

A good data source for enhancing sports performance and injury prediction should provide unique, sport-specific information about each athlete in the target sport area. The rapid development of sports evaluation bring us a few promising candidates:  
One promising data source comes from the rapidly developing field of wearable technology, which has impacted various aspects of life, including detecting potential falls~\cite{qu2015evaluation, qu2016real}, alerting users to abnormal heart rates, analyzing sleep quality~\cite{sathyanarayana2016sleep}, and tracking workout progress;  Another valuable data source is the growing use of athlete self-report measures (ASRM), which has shown great potential in enhancing athletic performance~\cite{saw2017athlete}; Videos of athletes' training sessions and matches offer valuable insights into an athlete's unique style and techniques. Video analysis has been widely used to enhance performance across various sports, including basketball, swimming, and tennis. 

 These data sources offer rich, objective, and subjective data that can be leveraged to improve predictions of performance and injury risk in tennis.  This motivates us to develop a powerful performance and injury prediction tool which is capable of capture unique information from all the promising diverse data sources and give detailed, meaningful predictions and suggestions for tennis players.
  The challenge is also obvious: the more data sources are considered, the complexity of developing such a prediction model will increase dramatically. 
  To use the wearable device data and self-reported survey, a more novel design of a prediction framework is necessary. Further, when the video is considered, how to incorporate the video data into the framework to provide meaningful insight of the athletes makes the design even more challenging. 
 
To this end, this paper proposes a \emph{multimodal} \textbf{P}redictive \textbf{A}thlete \textbf{R}eadiness framework for \textbf{T}ennis ({\THESYSTEM}) that predicts the wellness of tennis athletes considering both the physical capabilities as well as injury probabilities. The athlete's physical capability, as predicted by the regression model, directly impacts their injury risk.
A lower physical capability, potentially due to fatigue or inadequate recovery, can significantly increase the susceptibility to injury. 
This framework utilizes a variety of machine learning algorithms to develop a predictive model that integrates data from multiple sources, including {videos}, physiological data, workout details, sleep information, and self-reported survey responses from collegiate tennis players. 
Specifically, our contributions are:

\begin{enumerate}
    \item We collected diverse sources of data from $9$ (male/female) collegiate tennis athletes including both video-type data such as match videos, and textual-type data such as physiological data, workout details and sleep information from wearable devices, self-reported survey responses and regular vertical jump data recorded by professional physicians. Each source of data has provides its own unique information in performance and injury prediction for tennis players.
    \item We developed a \emph{mutlimodal} prediction framework {\THESYSTEM} which
    predicts the performance capacity and injury susceptibility of tennis players via our defined Athlete Readiness Score (ARS) along with the body areas with high injury risks. Our ARS is learned by supervised learning over multiple factors: physical capability prediction using deep learning models MLP and LSTM; injury risk probability using XGBoost Classifier; overall wellness prediction using XGBoost Regressor; playing style analysis using
    computer vision and motion analysis techniques. The playing style analysis handles the video-type data while the others focus on textual-type data.    
\end{enumerate}
This paper is organized as follows: 
Section~\ref{arch} first gives a overview of the architecture of our framework. Section~\ref{data} focuses on data collection and data preprocessing and Feature Engineering. Section~\ref{model} shows four models for overall wellness prediction, injury risk classification, physical capability prediction and playing style analysis. Section~\ref{integration} shows the details of generating ARS by the integration of these models' results. Section~\ref{relatedwork} shows the related work in sports performance and injury prediction. We conclude our paper and discuss about the future work in Section~\ref{conclusion}.

\section{System Architecture}
\label{arch}

\begin{figure*}[!h]
\centering
\includegraphics[width=0.95\textwidth]{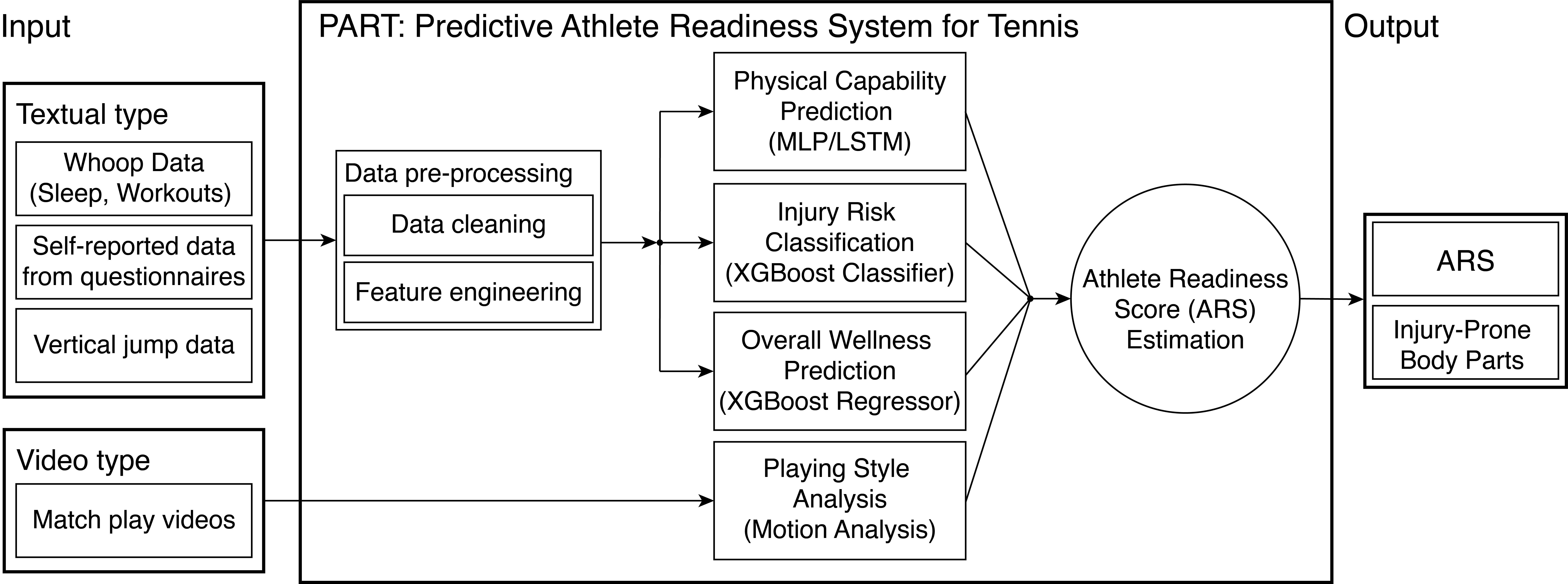}
\caption{System architecture for athlete wellness and injury risk prediction.}
\label{fig:system_architecture}
\end{figure*}

Building upon the collected and preprocessed data, we developed a comprehensive system architecture to predict athlete wellness and injury risk. The architecture integrates various machine learning models and facilitates seamless data flow from input to output.
Figure~\ref{fig:system_architecture} illustrates the overall architecture of {\THESYSTEM}. It consists of data input layers, preprocessing modules, four models learning the physical capability, injury risk, overall wellness and playing style of the tennis athletes, and an integration model to generate the outputs.  This modular design allows for flexibility and scalability in processing diverse data types and integrating new models as needed. Our framework begins with data collection from four primary sources: WHOOP wearable devices providing sleep and workout data, self-reported data from questionnaires, and vertical jump data from profession physicians and videos of match play. The first three are textual-type data with a rigorous preprocessing phase, including data cleaning and feature engineering.

At the core of {\THESYSTEM} are four specialized predictive models: 1) A Physical Capability Prediction Model utilizes MLP and LSTM architectures to forecast an athlete's physical readiness based on temporal sequences of data; 2) An Injury Risk Classification Model employs an XGBoost Classifier to assess the probability of injury, leveraging both physiological and self-reported data; 3) An Overall Wellness Prediction Model relies on an XGBoost Regressor to evaluate the athlete's general well-being by integrating multiple data features.
4) A playing style model to predict the athlete's style based on match play videos. We choose XGBoost for its the best performance compared to other models, details in Section~\ref{model}.

The outputs from these four models are synthesized in our integration model to calculate an Athlete Readiness Score (ARS), providing a holistic assessment of the athlete's condition. 
The ARS is a composite score ranging from 0 to 100, where higher scores indicate better overall readiness for performance. For instance, an ARS of 90 or above suggests that the athlete is in optimal condition for high-intensity training or competition, while a score below 60 may indicate a need for additional rest or recovery.
In the integration model, ARS will be learned by supervised learning. Details are in Section~\ref{model}.
Besides, our framework also provides the injury area prediction, which comes from the injury risk model.

\section{Data Collection and Preprocessing}
\label{data}

\subsection{Data Collection}
Data were collected from nine collegiate tennis players (Male: 5, Female: 4; Age: 20.3 ± 1.5 years; Height: 175.2 ± 8.7 cm; Weight: 68.9 ± 9.2 kg) over a period of 16 weeks. 
All participants were informed about the study procedures and provided written consent.

\begin{figure*}[h]
\centering
\begin{subfigure}{0.30\textwidth}
\includegraphics[width=\linewidth]{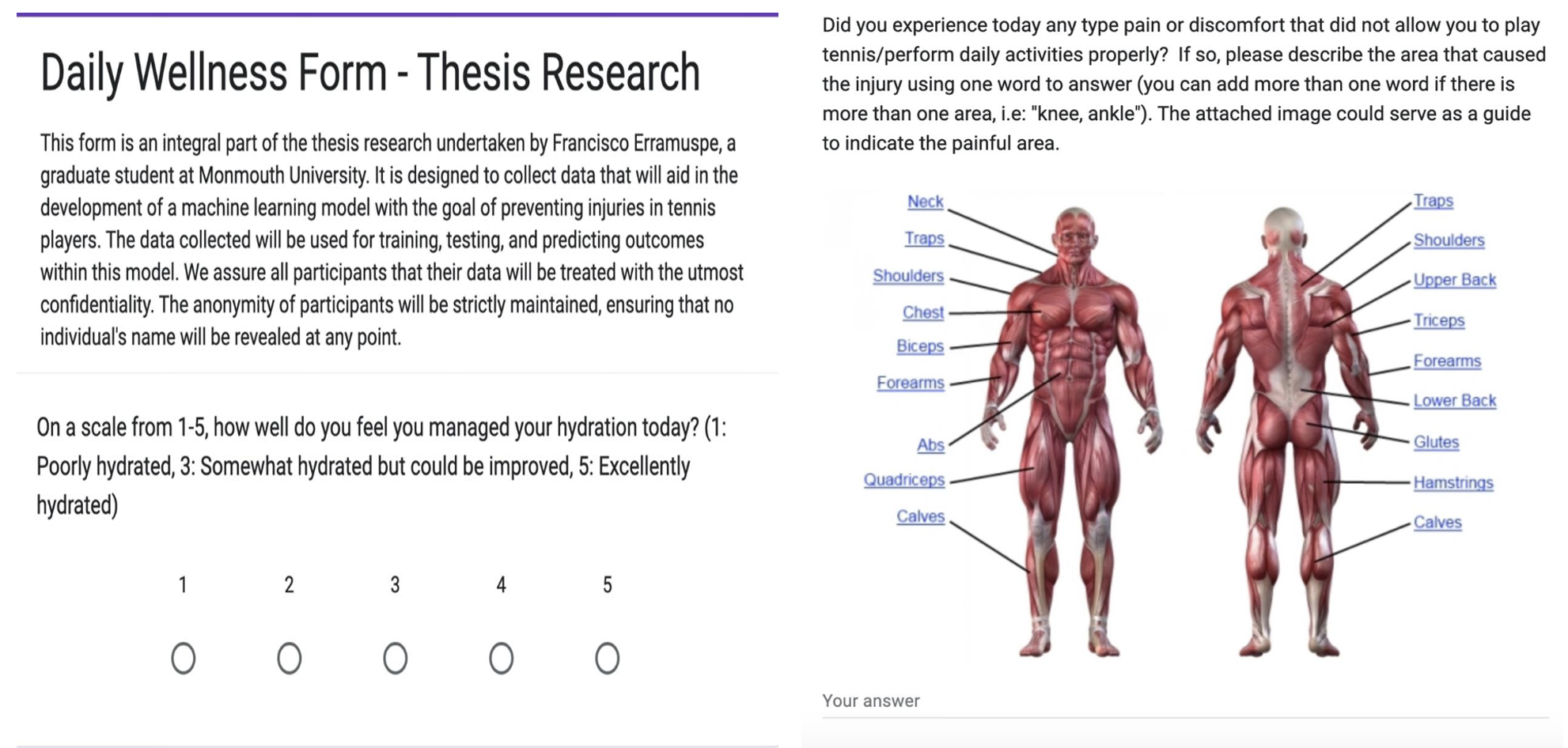}
\caption{Questionnaire}
\end{subfigure}
\hfill
\begin{subfigure}{0.24\textwidth}
\includegraphics[width=\linewidth]{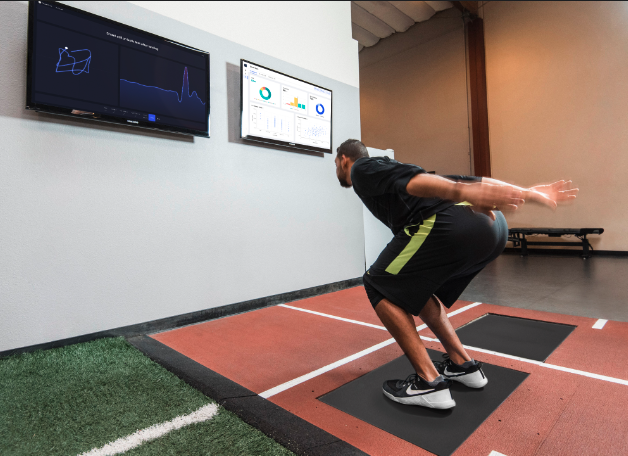}
\caption{Sparta jump}
\end{subfigure}
\hfill
\begin{subfigure}{0.40\textwidth}
\includegraphics[width=\linewidth]{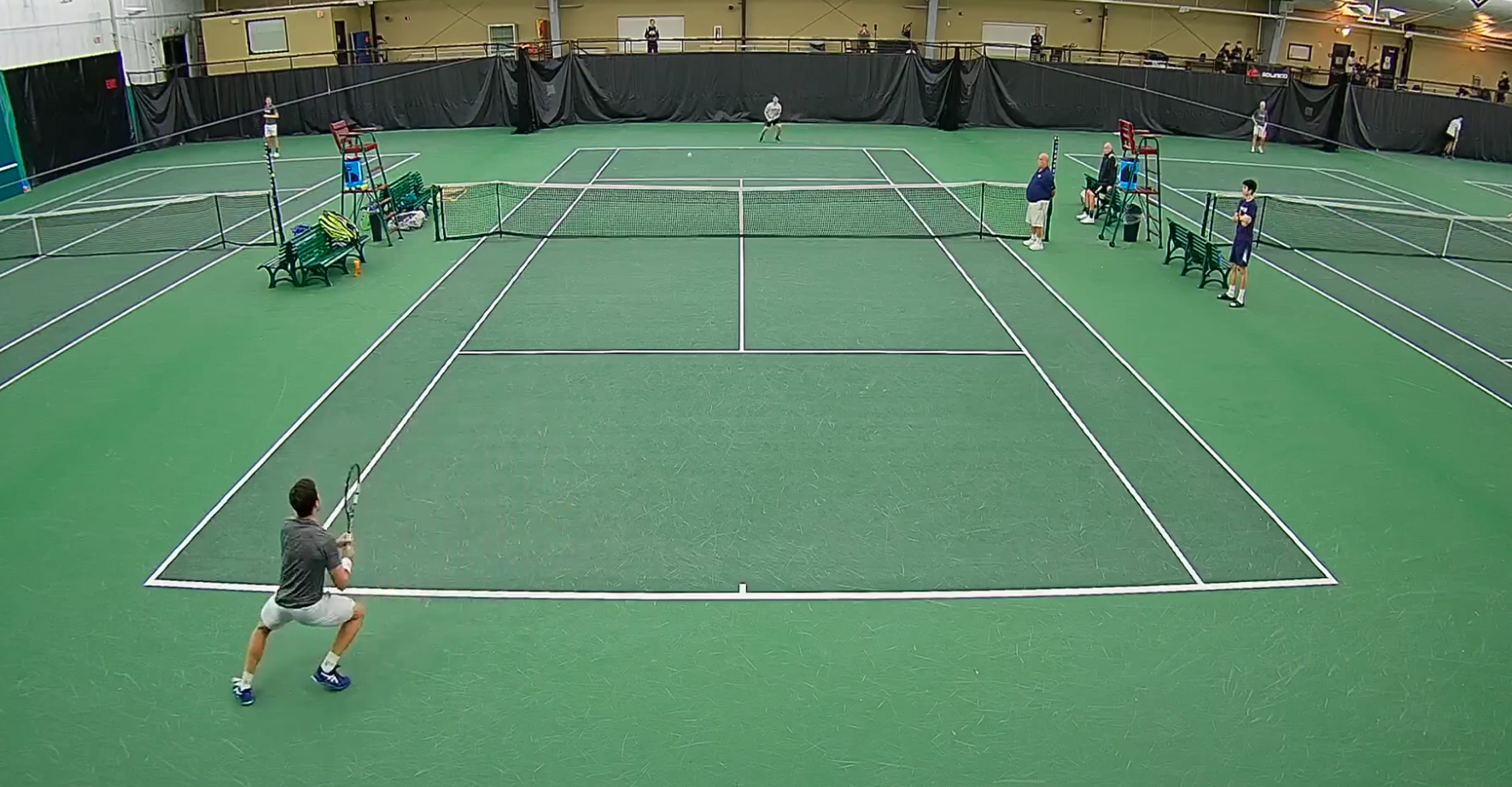}
\caption{Match play video}
\end{subfigure}
\caption{Illustrative input data types used in the framework: (a) online questionnaire about various aspects of athletes, (b) vertical jump data from Sparta testing to evaluate physical performance and readiness, and (c) match play videos for motion analysis and playing style assessment. WHOOP wearable metrics are also used by the framework but are not pictured here.}
\label{fig:questionnaire}
\end{figure*}

Data were gathered from four distinct sources. Figure~\ref{fig:questionnaire} illustrates three of them: daily online \emph{questionnaires}, vertical jump testing, and match play videos. In addition, WHOOP wearable devices recorded objective sleep data, continuous physiological data, and detailed workout data encompassing training sessions and matches. We collected vertical jump data using a standardized weekly or bi-weekly Sparta testing protocol as shown in Figure~\ref{fig:questionnaire}(b), and match play videos as shown in Figure~\ref{fig:questionnaire}(c). \wq{The number of attributes we collected is $85$ and the number of entries is $82804$.}

\subsection{Data Preprocessing and Feature Engineering}

The data preprocessing involves several critical steps to ensure data quality and suitability for model training. These steps include standardizing missing values, converting data types, imputing missing values, creating the 'Injury Risk' target variable, selecting relevant features, handling multicollinearity, and scaling numerical features.
\wq{
To handle the missing values, numerical features Resting heart rate (bpm) will be imputed using the mean or median, depending on the distribution, and categorical columns, such as Activity name  will be imputed with the most frequent value (mode).
For time-series data, missing timestamps such as the Workout start time will be forward-filled to maintain continuity.
}

Feature engineering enhances the model's ability to capture complex patterns and relationships within the data. 
We use the following feature engineering techniques applied in this study. To address skewed distributions and satisfy linear regression assumptions, we applied Logarithmic Transformations to variables such as ``Heart Rate Variability (ms)''to help normalize the data distribution. Interaction Terms between key variables like ``Activity Strain'' and ``Asleep duration (min)'' are created to capture combined effects on recovery, allowing models to account for the interplay between different factors affecting an athlete's wellness.
To to capture non-linear relationships between predictors and the target variable, we generated Polynomial Features, which expands the feature space. 
Lastly, we performed Categorical Variable Encoding on variables such as ``Activity Name'' using one-hot encoding. This process converts categorical data into a suitable numerical format, ensuring that all available information is utilized effectively.

\section{Model Development}
\label{model}
We developed different models for textual-type data and video-type data to learn the wellness, injury risk, injury area, physical capability and playing style of the tennis athletes.

\subsection{Overall Wellness Prediction (Regression Models)}

The objective of Overall Wellness Prediction is to predict the Recovery Score (\%), which ranges from 0 to 100 and indicates how well an athlete has recovered from physical exertion. 
The key physiological metrics used are:

\begin{itemize}
    \item \textbf{Heart Rate Variability (HRV)}: Measures the variation in time between heartbeats, showing stress and recovery.
    \item \textbf{Resting Heart Rate (RHR)}: A lower resting heart rate suggests better fitness and recovery.
    \item \textbf{Sleep Efficiency}: The percentage of time spent asleep while in bed, indicating recovery quality.
    \item \textbf{Activity Strain}: A measure of the intensity of physical activity performed by the athlete.
\end{itemize}

We utilized regression models including Linear Regression (with various enhancements such as feature transformation and interaction terms), Polynomial Regression, Lasso Regression, and XGBoost Regressor. These models aimed to predict the Recovery Score (\%), a key indicator of athlete readiness.

We used Mean Absolute Error (MAE), Root Mean Squared Error (RMSE), and R-squared (R²) Score. Results show that the XGBoost Regressor performed the best, achieving an MAE of 3.82 and an R² of 0.838, indicating that the model explains 83.8\% of the variance in recovery scores. Table \ref{tab:regression_results} presents the performance metrics for these models.

\begin{table}[h]
\caption{Regression Performance(Overall Wellness)}
\label{tab:regression_results}
\centering
\begin{tabular}{lccc}
\hline
\textbf{Model} & \textbf{MAE} & \textbf{RMSE} & \textbf{R\textsuperscript{2} Score} \\
\hline
Baseline Linear Regression & 14.05 & 16.84 & 0.009 \\
Linear Regression (Feature Transform) & 11.32 & 14.27 & 0.215 \\
Linear Regression (Interaction Terms) & 10.47 & 13.58 & 0.310 \\
Polynomial Regression & 8.23 & 11.40 & 0.537 \\
Lasso Regression & 9.76 & 12.85 & 0.402 \\
XGBoost Regressor & 3.82 & 6.81 & 0.838 \\
\hline
\end{tabular}
\end{table}
\subsection{Injury Risk Classification (Classification Models)}

The goal of Injury Risk Classification is to predict whether an athlete is at High Risk or Low Risk of injury within a specific time frame. The classification was based on physiological metrics such as Activity Strain, HRV, and Resting Heart Rate, as well as self-reported survey responses.

We implemented Logistic Regression, XGBoost Classifier, Decision Tree Classifier, and Random Forest Classifier. These models were tasked with predicting the likelihood of an athlete sustaining an injury based on the collected data.
The models were not only used to predict whether an athlete would get injured, but also where and how the injury might occur, enabling personalized interventions:

\begin{itemize}
    \item \textbf{Upper Body Injuries:} arms, shoulders, and upper torso.
    \item \textbf{Lower Body Injuries:} legs, knees, and lower torso.
    \item \textbf{Left Side / Right Side:} Injuries classified based on which side of the body was affected.
    \item \textbf{Upper Left, Upper Right, Lower Left, Lower Right:} Quadrant-based classification to specify injury regions.
\end{itemize}

We employed Accuracy, F1 Score, and AUC-ROC for evaluation. We compared various classification models for overall injury risk prediction. 
Table \ref{tab:classification_results} summarizes their performance.
\wq{ Our results indicate that the XGBoost Classifier outperformed other models, achieving the highest F1 score and AUC-ROC, demonstrating its ability to correctly identify injury risks. In contrast, Logistic Regression had an F1 score of 0, highlighting its failure to predict injury cases due to the imbalance in the dataset and its limitations in capturing complex patterns. }

\begin{table}[h]
\caption{Classification Performance(Injury Risk Prediction)}
\label{tab:classification_results}
\centering
\begin{tabular}{lccc}
\hline
\textbf{Model} & \textbf{AUC-ROC} & \textbf{Accuracy} & \textbf{F1 Score} \\
\hline
Logistic Regression & 0.546 & 0.849 & 0.000 \\
XGBoost Classifier  & 0.695 & 0.852 & 0.100 \\
Decision Tree Classifier & 0.622 & 0.840 & 0.075 \\
Random Forest Classifier & 0.678 & 0.850 & 0.092 \\
\hline
\end{tabular}
\end{table}
 Table \ref{tab:injury_type_prediction} shows the performance of XGBoost models for upper-body and lower-body injury predictions.
 
\begin{table}[h]
\caption{Performance of Injury Type Prediction Models}
\label{tab:injury_type_prediction}
\centering
\begin{tabular}{lccc}
\hline
\textbf{Injury Type} & \textbf{AUC-ROC} & \textbf{Accuracy} & \textbf{F1 Score} \\
\hline
Upper Body & 0.712 & 0.865 & 0.132 \\
Lower Body & 0.703 & 0.858 & 0.118 \\
\hline
\end{tabular}
\end{table}

\subsection{Physical Capability Prediction}

We developed models using both Multi-Layer Perceptron (MLP) and Long Short-Term Memory (LSTM) architectures to predict the athlete's physical capability percentage based on a 7-day sequence of physiological and activity metrics.  \wq{We choose one week as the time interval for LSTM due to its best performance.}

The athlete's physical capability was quantified using the Physical Capability Score (PCS), calculated through a weighted formula that combines various physiological and subjective metrics:

\begin{align}
\text{PCS} = &\ 0.25 \times \mathrm{RS} + 0.20 \times \mathrm{SE} + 0.15 \times \mathrm{HRV}_{\mathrm{score}} \nonumber \\
             &\ + 0.15 \times \mathrm{SS} + 0.15 \times \mathrm{QS} + 0.10 \times \mathrm{VJP}_{\mathrm{score}} \nonumber
\end{align}

\begin{table}[h]
\caption{Data source of notations used in PCS.}
    \centering
    \begin{tabular}{|l|c|}
    \hline
       Notation  & Source \\
       \hline
       Reactive Strength (RS) & Jump data \\
       \hline
       Sleep Efficiency (SE) & Whoop \\
       \hline
       Heart Rate Variability Score (HRV$_{\mathrm{score}}$) & Whoop \\
              \hline
       Sleep Score (SS) & Whoop \\
       \hline
       Quality Score (QS) & Questionnaire \\
       \hline
       Vertical Jump Performance Score (VJP$_{\mathrm{score}}$) & Jump data \\
       \hline
    \end{tabular}
    \label{tab:pcs}
\end{table}

  We show the sources of used notations in PCS in Table~\ref{tab:pcs}. {Reactive Strength (RS)} measures an athlete's ability to quickly transition from an eccentric to a concentric muscle contraction.
{Sleep Score (SS)} is a composite metric that evaluates overall sleep quality based on factors like sleep duration, sleep efficiency, disturbances, and sleep stages. Higher scores indicate better sleep quality. {Quality Score (QS)} is a subjective assessment of the athlete's perceived wellness and readiness, often derived from self-reported questionnaires covering aspects like mood, fatigue, and muscle soreness.{Vertical Jump Performance Score (VJP$_{\mathrm{score}}$)} is an objective measure of lower-body explosive power, assessed through vertical jump tests. It is critical for athletic performance in sports requiring jumping ability.

Our predictive models utilize the PCS calculated from historical data over a 7-day period to forecast the athlete's physical capability percentage. This approach allows us to account for recent trends in physiological state and activity levels, enhancing the accuracy of our predictions. \wq{We choose 80-20 split for training and validation. }
Table~\ref{tab:neural_network_performance} shows the results.
\begin{table}[h]
\caption{Performance (Physical Capability Prediction)}
\label{tab:neural_network_performance}
\centering
\begin{tabular}{lcc}
\hline
\textbf{Model} & \textbf{Training Loss} & \textbf{Validation Loss} \\
\hline
MLP & 0.0388 & 0.6391 \\
LSTM & 0.0360 & 0.6059 \\
\hline
\end{tabular}
\end{table}

\subsection{Playing Style Analysis}
To enhance our understanding of individual playing patterns and their impact on athlete wellness, we implemented an automated video analysis system to track and analyze match play. The system calculates the average number of shots per point (ASP) for each player, \wq{which considers all the points played in the video,} providing valuable insights into playing style and physical exertion patterns. Based on the ASP metric, players are classified into four distinct categories:

\begin{itemize}
    \item \textbf{Highly Aggressive} (ASP $\leq$ 5): Players who consistently end points quickly through powerful shots
    \item \textbf{Aggressive} (5 $<$ ASP $\leq$ 8): Players who maintain an attacking style while showing more rally tolerance
    \item \textbf{Neutral} (8 $<$ ASP $\leq$ 12): Players who balance aggressive and defensive play
    \item \textbf{Defensive} (ASP $>$ 12): Players who primarily rely on consistency and extended rallies
\end{itemize}

\section{Model Integration and Output Generation}
\label{integration}

\subsection{Athlete Readiness Score (ARS) Labelling}
\label{sec:ars_labelling}

To enable supervised learning for determining the optimal weights in the Athlete Readiness Score (ARS), we first established meaningful labels for the ARS based on expert evaluations. Coaches and sports scientists assessed each athlete's overall readiness using a comprehensive scale from 0 to 100. This assessment considered various factors, including physical performance metrics, recovery status, training load, and subjective wellness reports.

The labelling process involved the following steps:

\begin{enumerate}
    \item \textbf{Data Collection}: Gathering objective metrics (e.g., physiological data from wearables, performance statistics) and subjective reports (e.g., self-reported wellness, stress levels).
    \item \textbf{Expert Evaluation}: Coaches reviewed the collected data alongside their observations to assign an ARS label to each athlete for each time point.
    \item \textbf{Consensus Building}: To minimize individual bias, multiple experts provided assessments, and the final ARS label was obtained by averaging their scores.
    \item \textbf{Data Alignment}: The labelled ARS data was aligned with the corresponding features for supervised learning.
\end{enumerate}

This labelled dataset provided the ground truth necessary for training our models to predict ARS.

\subsection{Determining Weights via Supervised Learning}
\label{sec:weight_determination}

We aimed to let the model determine the optimal weights \( w_1, w_2, w_3 \) in the ARS formula through supervised learning. The ARS is computed as follows:

\vspace{-1em}
\begin{align}
\text{ARS}_{\text{initial}} = &\ w_1 \times \text{Overall Wellness Score} \nonumber \\ &\ + w_2 \times (1 - \text{Injury Risk Probability}) \nonumber \\
&\ + w_3 \times \text{Physical Capability Score}  \nonumber
\end{align}
Subject to the constraint:
$w_1 + w_2 + w_3 = 1$.

The weight determination process contains five stages: Feature Preparation, Model Selection, Model Training, Weight Extraction, Validation. We normalized the three component scores for each athlete in Feature Preparation. In the model selection, we chose Multiple Linear Regression due to its interpretability and ability to provide direct coefficients as weights. The training process uses the labelled ARS. In validation stage, we evaluated model performance using metrics such as R-squared and Mean Squared Error and conducted residual analysis to verify model assumptions.

\subsection{Incorporating Video Analysis Results}
\label{sec:video_incorporation}

The playing style from the video-type data serves as a important weighting factor in our Athlete Readiness Score (ARS) calculations, with higher weights assigned to recovery metrics for defensive players who typically experience greater physical demands during matches. The ASP metric is integrated into our framework using the following formula:
\vspace{-0.5em}
\begin{equation}
w_{style} = \begin{cases}
1.0 & \text{if ASP} \leq 5 \\
1.2 & \text{if } 5 < \text{ASP} \leq 8 \\
1.4 & \text{if } 8 < \text{ASP} \leq 12 \\
1.6 & \text{if ASP} > 12
\end{cases}
\end{equation}
The final ARS is then adjusted based on the player's style classification derived from video analysis:

\begin{equation}
ARS_{\text{final}} = \frac{ARS_{\text{initial}}}{w_{\text{style}}}
\end{equation}

This adjustment ensures that our wellness predictions account for the varying physical demands associated with different playing styles, providing more accurate and personalized assessments of athlete readiness and injury risk.

\subsection{Code Availability}

The code for data preprocessing, model training, and evaluation is available in a public GitHub repository at \url{https://github.com/franciscoerramuspe/masters_thesis}. The repository includes all scripts necessary to replicate the results discussed in this paper, as well as detailed instructions for setting up the environment and running the experiments.

\section{Related Work}
\label{relatedwork}

\textit{Integration of Wearable Technology and Self-Reported Data}

The integration of wearable technology data with self-reported metrics has been a recurring theme in sports analytics research. 
{
It has been reported that self-reported data can effectively provide new insights into the interactions between competition-related stressors experienced by professional athletes and their susceptibility to illness~\cite{thornton2016predicting}.
The wearable fitness band, Whoop, has been widely adopted in college athletics, including sports such as softball, women's lacrosse, baseball~\cite{romano2023examination}, and wrestling~\cite{gerardi2023exploring}.}
In the college basketball studies~\cite{taber2024,zhao2022}, the value of combining objective physiological data with subjective measures such as stress and recovery questionnaires is revealed to enhance the predictive power of ML models. 

\textit{Time-Series Data}
Recent advancements in deep learning have shown promise in capturing temporal dependencies and complex nonlinear relationships in time-series data.
{In sports like basketball~\cite{taber2024}, volleyball~\cite{de2022personalized}, tennis~\cite{myers2020acute,moreno2021association}, and soccer~\cite{nassis2023review}, data is typically collected over seasons, aligning with the characteristics of time series data.}
 Our implementation of MLP and LSTM models aims to leverage these strengths, enabling the prediction of an athlete's physical capability and readiness based on their physiological and activity metrics over time. 
 This approach not only complements traditional ML models but also addresses the temporal dynamics that are critical in understanding athlete performance and injury risk.

\textit{Specific to Tennis} 
There are several works focusing on college basketball athletes using different machine learning models such as XGBoost~\cite{taber2024,huang2022novel},  neural networks~\cite{zhao2022} for injury assessment. At the same time,  tennis studies often focus more on qualitative insights, such as player feedback and expert evaluations of technique. The research
~\cite{moreno2021association} studied the association and predictive ability of several markers of internal workload on risk of injury in high-performance junior tennis players and shows that a high acute workload is one factor associated with injury. 
The investigation~\cite{myers2020acute} indicates an acute increase in
load was associated with increased injury risk after the analysis of the workload and self-reported injuries in junior tennis players.
The study~\cite{gescheit2018modelling} studies the injury facts among elite tennis players by examining the epidemiology and in-event treatment frequency of injury
at the 2011-2016 Australian Open tournaments. The data includes sex, injury region, and type and is reported as frequencies per 10,000 game exposures.
Also, features such as player characteristics, injury histories, recent tournament schedules~\cite{liu2023sports}, and the physical components like “upper body power”, “lower body power”, “speed”, and “agility” ~\cite{kramer2017prediction} are applied to develop a predictive model capable of estimating an individual’s probability of getting injured in the future by Regression analyses. 
To predict the match outcome, the work ~\cite{tennisMatch} used a supervised machine learning approach that uses historical player performance across a wide variety of statistics, including logistic regression and artificial
neural networks evaluated on a test set of 6315 ATP matches played in one year.
According to the study~\cite{sampaio2024applications}, 
the reliance on subjective data limits the ability to capture the full complexity of performance and injury risks in tennis, highlighting the need for more comprehensive, data-driven approaches, which is the motivation of our framework.

\section{Conclusion and Future works}
\label{conclusion}
We collect data from nine college tennis athletes in one semester by means of online questionnaires, wearable devices, professional testing protocol, and video recording. Then we present a novel multimodal framework {\THESYSTEM} providing robust wellness and injury prediction for tennis athletes by integrating various sources of data using varied machine learning techniques. {\THESYSTEM} could predict both the condition of tennis athletes precisely as well as the injury-prone body parts.

There are a few future directions we plan to improve our work: 1) We plan to provide more subtle injury-prone body part prediction beyond the current left, right, upper and lower body results. 2) We plan to develop a more robust approach to incorporating the video data result in our prediction. 3) We will implement the framework as an Intelligent Tennis Coach software and promote it to tennis players at all levels.

\bibliographystyle{IEEEtran}
\bibliography{main}

\end{document}